# Local Structure Discovery in Bayesian Networks

**Teppo Niinimäki**
Helsinki Institute for Information Technology HIIT
Department of Computer Science
University of Helsinki, Finland
`teppo.niinimaki@cs.helsinki.fi`

**Pekka Parviainen**
CSC and Scilifelab
Royal Institute of Technology (KTH)
Stockholm, Sweden
`pekkapa@kth.se`

## Abstract

Learning a Bayesian network structure from data is an NP-hard problem and thus exact algorithms are feasible only for small data sets. Therefore, network structures for larger networks are usually learned with various heuristics. Another approach to scaling up the structure learning is local learning. In local learning, the modeler has one or more target variables that are of special interest; he wants to learn the structure near the target variables and is not interested in the rest of the variables. In this paper, we present a score-based local learning algorithm called SLL. We conjecture that our algorithm is theoretically sound in the sense that it is optimal in the limit of large sample size. Empirical results suggest that SLL is competitive when compared to the constraint-based HITON algorithm. We also study the prospects of constructing the network structure for the whole node set based on local results by presenting two algorithms and comparing them to several heuristics.

## 1 INTRODUCTION

A Bayesian network is a representation of a joint probability distribution. It consists of two parts: the structure and the parameters. The structure of a Bayesian network is represented by a directed acyclic graph (DAG) that expresses the conditional independence relations among variables and the parameters determine the local conditional distributions for each variable.

Structure learning in Bayesian networks, that is, finding a DAG that fits to the data, has drawn lots of interest in recent years. There are two rather distinct approaches to structure learning. In score-based approach (Cooper and Herskovits, 1992; Heckerman et al., 1995) each DAG gets a score based on how well it fits to the data and the goal is to find the DAG that maximizes the score. On the other hand, the constraint-based approach (Pearl, 2000; Spirtes et al., 2000) is based on testing conditional independencies among variables and constructing a DAG that represents these relations. Both approaches are compatible in the sense that (under some general assumptions), no matter which approach one uses, exact algorithms converge to the same DAG when the number of samples tends to infinity.

Structure learning in Bayesian networks is NP-hard (Chickering, 1996; Chickering et al., 2004). Due to the NP-hardness of the problem, it is unlikely that there are exact algorithms that run in polynomial time. Indeed, the fastest exact algorithms (Silander and Myllymäki, 2006) run in exponential time. Although there have been several attempts to scale up the exact algorithms (de Campos et al., 2009; Parviainen and Koivisto, 2009; Jaakkola et al., 2010; Cussens, 2011; Malone et al., 2011), these methods quickly become infeasible with larger datasets.

As the exact algorithms do not work for large data sets, one often has to resort to heuristics. A common strategy is to use different variants of greedy search. An alternative approach to the problem is local learning. When one has a large data set, often it is a case that all variables are not equally interesting. One might have one or a handful of *target variables* and the goal is to learn the structure only in the vicinity of the targets. An example for a problem where local learning is feasible is prediction. If we know the Markov blanket of a target variable, that is, its parents, its children and the parents of its children, the remaining variables do not give any more information. Thus, if it is known that we are going to predict the values of only a handful of variables, we can just learn their Markov blankets and ignore the structure of the rest of the network.

Local learning is not a new approach. There have been several studies especially on constraint-based lo-cal learning (Nägele et al., 2007; Aliferis et al., 2010a,b). These studies have shown that local learning can be a powerful tool in practice. In this paper we study the prospects of the score-based local learning of Bayesian network structures. We present a *score-based local learning algorithm (SLL)* that is a variant of the generalized local learning (GLL) framework (Aliferis et al., 2010a). Assuming a consistent scoring criterion is used, we conjecture that our algorithm is theoretically sound in the same sense as greedy equivalence search (GES) (Chickering, 2002) and HITON (Aliferis et al., 2010a), that is, it is guaranteed to find the true Markov blanket of the target node when the sample size tends to infinity.

Optimality guarantees in the limit or lack thereof do not tell anything about results with finite sample size. Thus, we conducted experiments and compared our algorithm to other heuristics. Based on the experiments, we found our algorithm promising.

## 2 PRELIMINARIES

### 2.1 BAYESIAN NETWORKS

The structure of a Bayesian network is represented by a *directed acyclic graph (DAG)*. Formally, a DAG is a pair $(N, A)$ where $N$ is the node set and $A$ is the arc set. If there is an arc from $u$ to $v$, that is, $uv \in A$, we say that $u$ is a *parent* of $v$ and $v$ is a *child* of $u$. If $v$ is either a parent or a child of $u$, they are said to be *neighbors*. Further, if there is a directed path from $u$ to $v$ in $A$ then $v$ is a *descendant* of $u$. Nodes $v$ and $u$ are are said to be *spouses* of each other if they have a common child and there is no arc between $v$ and $u$. We denote the set of the parents of a node $v$ in $A$ by $A_v$. Further, we use $H^A(v)$ and $S^A(v)$ to denote the sets of the neighbors and spouses of $v$ in $A$, respectively. When the node set is clear from the context, we identify a DAG by its arc set $A$. The cardinality of $N$ is denoted by $n$.

Each node $v$ corresponds to a random variable and the DAG expresses conditional independence assumptions between variables. Random variables $u$ and $v$ are said to be conditionally independent in distribution $p$ given a set $S$ of random variables if $p(u, v|S) = p(u|S)p(v|S)$. A DAG *contains* or *represents* a joint distribution of the random variables if the joint distribution satisfies the *local Markov condition*, that is, every variable is conditionally independent of its non-descendants given its parents. Such a distribution can be specified using local *conditional probability distributions (CPD)* which specify the distribution of a random variable given its parents $A_v$. CPDs are usually taken from a parameterized class of probability distributions, like discrete or Gaussian distributions. Thus, the CPD of variable $v$ is determined by its parameters $\theta_v$; the type and the number of parameters is specified by the particular class of probability distributions. The parameters of a Bayesian network are denoted by $\theta$ which consists of the parameters of each CPD. Finally, a Bayesian network is a pair $(A, \theta)$.

A distribution $p$ is said to be *faithful* to a DAG $A$ if all conditional independencies in $p$ are implied by $A$. We say that a DAG $A^p$ is a *perfect map* of a distribution $p$ if $A^p$ contains $p$ and $p$ is faithful to $A^p$. By $p[Z]$ we denote the marginal distribution of $p$ on set $Z$.

The conditional independencies implied by a DAG can be extracted using a d-separation criterion; this is equivalent to local Markov condition. The *skeleton* of a DAG $A$ is an undirected graph that is obtained by replacing all directed arcs $uv \in A$ with undirected edges between $u$ and $v$. A *path* in a DAG is a cycle-free sequence of edges in the corresponding skeleton. A node $v$ is a *head-to-head node* along a path if there are two consecutive arcs $uv$ and $wv$ on that path. Nodes $v$ and $u$ are *d-connected* by nodes $Z$ along a path from $v$ to $u$ if every head-to-head node along the path is in $Z$ or has a descendant in $Z$ and none of the other nodes along the path is in $Z$. Nodes $v$ and $u$ are *d-separated* by nodes $Z$ if they are not d-connected by $Z$ along any path from $v$ to $u$. If $A^p$ is a perfect map of $p$ then $v$ and $u$ are conditionally independent give $Z$ in $p$ if and only if $v$ and $u$ are d-separated by $Z$ in $A^p$. If $v$ and $u$ are conditionally independent in $p$ given $Z$ we use notation $v \perp\!\!\!\perp_p u | Z$.

The *Markov blanket* of node $v$ is the smallest node set $S$ such that $v$ is conditionally independent of all other nodes given $S$. The Markov blanket of node $v$ consists of the parents of $v$, the children of $v$, and the spouses of $v$.

Nodes $s$, $t$, and $u$ form a *v-structure* in a DAG if $s$ and $t$ are spouses and $u$ is their common child. A v-structure is denoted by $(s, u, t)$. Two DAGs are said to be *Markov equivalent* if they can contain the same set of distributions, or equivalently, imply the same set of conditional independence statements. It can be shown that two DAGs are Markov equivalent if and only if they have the same skeleton and same v-structures (Verma and Pearl, 1990).

### 2.2 CONSISTENT SCORING CRITERION

Score-based structure learning methods in Bayesian networks assign a score to each DAG based on how well the DAG fits to the data according to some statistical principle. A function $f(A, D)$ that assigns the score to a DAG $A$ based on the data $D$ is called a *scoring criterion*. A scoring criterion is *decomposable* if the score of a DAG is a sum of local scores that only depend on a node and its parents. A local score function is denoted by $f(v, A_v, D_v, D_{A_v})$, where $D_v$ and $D_{A_v}$ are the data on a node $v$ and a node set $A_v$, respectively. Now, a decomposable score for a DAG $A$ can be written as
$$f(A, D) = \sum_{v \in N} f(v, A_v, D_v, D_{A_v}).$$

Let the data $D$ consists of $m$ independent and identically distributed (i.i.d.) samples from a distribution $p$ on $N$. A scoring criterion $f$ is said to be *consistent* if in the limit as $m$ grows, the following two properties hold:

1. If $A'$ contains $p$ and $A$ does not, then $f(A', D) > f(A, D)$.

2. If $A'$ and $A$ both contain $p$, and $A'$ has fewer parameters, then $f(A', D) > f(A, D)$.

A related and in our case more useful property is local consistency. Let $A$ be any DAG and $A'$ be the DAG that results from adding the arc $uv$ to $A$. A scoring criterion $f$ is said to be *locally consistent* if in the limit as $m$ grows, the following two properties hold for all $u$ and $v$:

1. If $u \not\perp\!\!\!\perp_p v | A_v$, then $f(A', D) > f(A, D)$.

2. If $u \perp\!\!\!\perp_p v | A_v$, then $f(A', D) < f(A, D)$.

Intuitively, adding an arc that eliminates an independence constraint that does not hold in the data-generating distribution increases the score, and adding an arc that does not eliminate such a constraint decreases the score.

A scoring criterion is said to be *score equivalent* if two Markov equivalent DAGs always have the same score. With a score equivalent scoring criterion one can learn a structure of a Bayesian network up to an equivalence class but cannot distinguish the DAGs inside the class.

For example, commonly used BDeu score is locally consistent (Chickering, 2002) and score equivalent (Heckerman et al., 1995).

### 2.3 STRUCTURE DISCOVERY PROBLEM

The problem of structure discovery in Bayesian networks is to find, given data $D$, a DAG that in some sense is the best representation of the data. We call this global learning. In score-based approach the goodness of a DAG is measured by the score $f(A, D)$. In constraint-based approach one would like to find a DAG that is a perfect map of the data-generating distribution.

In the local learning problem the output is the neighbor and/or spouse sets for a target node, not a DAG. The goodness of the result can be measured by comparing learned neighbor and spouse sets to the corresponding sets in the $A^p$, the perfect map of the data-generating distribution. Note that this comparison can be made only if $A^p$ is known.

## 3 LOCAL LEARNING ALGORITHM

In this section, we present a score-based local learning algorithm algorithm (SLL) that finds the Markov blanket of a given target node. SLL works in two phases: First, one learns the neighbors of the target and then the rest of the Markov blanket. SLL is a score-based variant of the constraint-based local learning algorithm by Aliferis et al. (2010a,b); the main difference between SLL and the algorithm by Aliferis et al. is that they use independence tests to identify the neighbors and spouses whereas we recognize them from optimal Bayesian networks. In this section, we also analyze the behavior of the algorithm in the limit. We also consider constructing a Bayesian network for the whole node set using the local structures that have been found.

### 3.1 FINDING PARENTS AND CHILDREN

We start by learning the potential neighbors of a target node. The idea of the algorithm is as follows. The input of the algorithm consists of the data $D$ on the node set $N$ and a target node $t \in N$. During the execution of the algorithm, we update two sets: Set $O$ consists of the nodes that have not been analyzed yet and set $H$ consists of nodes that are currently considered as potential neighbors of $t$. In the beginning, set $H$ is empty and $O$ contains all nodes but $t$. After the initialization, nodes in $O$ are considered one by one: One learns an optimal DAG on the target, the current set of its potential neighbors and the new node under consideration. For the structure learning we use a subroutine OPTIMALNETWORK which returns the highest scoring DAG on a given node set. The subroutine can use the dynamic programming algorithm of Silander and Myllymäki (2006) or any other exact algorithm. After an optimal DAG is found, the nodes that are the neighbors of $t$ in that particular DAG form a new set of potential neighbors and the rest of the nodes are discarded. After all nodes are either discarded or in the set of potential neighbors, the set of the potential neighbors is returned.

The pseudocode of the above procedure is shown in Algorithm 1.

Next, we will show that if the data was generated from a distribution $p$, Algorithm 1 is guaranteed to find all correct neighbors, that is, all nodes that are neighbors of $t$ in $A^p$ will be included in $H(t)$ in the limit. The guarantee holds under the following assumptions.

**Algorithm 1** FINDPOTENTIALNEIGHBORS
**Input:** Data $D$ on node set $N$, a target node $t \in N$.
**Output:** The potential neighbors of $t$.
1: Initialization: $O \leftarrow N \setminus \{t\}$, $H(t) \leftarrow \emptyset$
2: **while** $O$ is nonempty **do**
3:    Choose $v \in O$
4:    $O \leftarrow O \setminus \{v\}$
5:    $Z \leftarrow \{t, v\} \cup H(t)$
6:    $A \leftarrow$ OPTIMALNETWORK$(Z, D_Z)$
7:    $H(t) \leftarrow H^A(t)$
8: **end while**
9: **return** $H(t)$

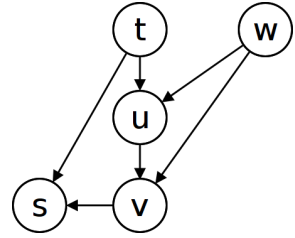

Figure 1: A DAG where non-adjacent nodes $t$ and $v$ are not conditionally independent given any subset of the neighbors of $t$.

**Assumption 1.** *The data $D$ consists of i.i.d. samples from a distribution $p$ that is faithful to a DAG $A^p$.*

**Assumption 2.** *The procedure OPTIMALNETWORK uses a locally consistent and score equivalent scoring criterion.*

**Lemma 3.** *Let $H^{A^p}(t)$ be the neighbors of the target $t$ in $A^p$. When Assumptions 1 and 2 hold, Algorithm 1 will return a set $H(t)$ such that $H^{A^p}(t) \subseteq H(t)$.*

*Proof.* If nodes $t$ and $v$ are not conditionally independent in $p$ given any set $X \subseteq N \setminus \{t,v\}$, then any Bayesian network that contains $p$ must have an arc between $t$ and $v$. Thus, if we have two networks $A$ and $A'$ on $Y \cup \{t,v\}$ such that $A'$ has an arc between $t$ and $v$ and $A$ has not and otherwise they are the same, then by the local consistency, $A'$ has higher score than $A$. Thus, the highest scoring network must have an arc between $t$ and $v$. As the above reasoning applies to every set $Y \subseteq N \setminus \{t,v\}$, $v$ will be in $H(t)$ when Algorithm 1 stops. □

Based on the previous lemma we know that if a node $v$ is a neighbor of a node $t$ in $A^p$, then it will be in $H(t)$. However, we have no guarantees that no nodes that are non-adjacent to $t$ in $A^p$ are included in $H(t)$. Indeed, Aliferis et al. (2010a) point out that if we have a perfect map $A^p$ described in Figure 1 an extra node might be added to $H(t)$. To see this, assume that $t$ is our target and its true neighbors $u$ and $s$ are in $H(t)$ and $w$ has been discarded. Then $t \not\perp\!\!\!\perp_p v | X$ for all $X \subseteq \{u, s\}$ and thus, by local consistency, adding an arc between $t$ and $v$ always increases the score and therefore it is always included in the optimal DAG.

The neighbor relation is symmetric: if $v$ is a child of $t$ then $t$ is a parent of $v$. This allows us to try to remove extra nodes from $H(t)$ using a simple symmetry correction in similar fashion as Aliferis et al. (2010a): $t$ and $v$ are neighbors in $A^p$ only if both $v$ is a potential neighbor of $t$ and $t$ is a potential neighbor of $v$. Algorithm 2 uses symmetry correction to find the neighbors of a target node.

**Algorithm 2** FINDNEIGHBORS
**Input:** Data $D$ on node set $N$, a target node $t \in N$.
**Output:** The neighbors of $t$.
1: $H^*(t) \leftarrow$ FINDPOTENTIALNEIGHBORS$(D, t)$
2: **for all** $v \in H^*(t)$ **do**
3:    $H(v) \leftarrow$ FINDPOTENTIALNEIGHBORS$(D, v)$
4:    **if** $t \notin H(v)$ **then**
5:       $H^*(t) \leftarrow H^*(t) \setminus \{v\}$
6:    **end if**
7: **end for**
8: **return** $H^*(t)$

To analyze the optimality of the algorithm, we use the below lemma which shows that if $t$ and $v$ are not neighbors in $A^p$ but they are dependent given every subset of the neighbors of $t$ then they are conditionally independent given some subset of the neighbors of $v$. Formally, let $E^{A^p}(t)$ be the set consisting of the nodes $v \in N \setminus \{t\}$ such that $t \not\perp\!\!\!\perp_p v | X$ for all $X \subseteq H^{A^p}(t)$.

**Lemma 4** (Aliferis et al. (2010a), Lemma 2). *If $v \in E^{A^p}(t) \setminus H^{A^p}(t)$, then $t \notin E^{A^p}(v) \setminus H^{A^p}(v)$.*

However, it is not clear whether the output of Algorithm 1 is a subset of the the nodes that are conditionally dependent on $t$ given any subset of the neighbors of $t$. As we are not aware of any counterexamples, we conjecture the follows.

**Conjecture 5.** *Let $H(t)$ be the output of Algorithm 1. Then $H(t) \subseteq E^{A^p}(t)$.*

The next lemma shows the optimality of Algorithm 2 in the limit assuming Conjecture 5 holds.

**Lemma 6.** *Let $H^{A^p}(t)$ be the neighbors of the target $t$ in $A^p$. When Assumptions 1 and 2 and Conjecture 5*

hold, Algorithm 2 will return a set $H^*(t)$ such that $H^*(t) = H^{A^p}(t)$.

*Proof.* First, let us prove that $H^{A^p}(t) \subseteq H^*(t)$. By Lemma 3, $H^{A^p}(t) \subseteq H(t)$ and $H^{A^p}(v) \subseteq H(v)$ for all $v$. Thus, if $v \in H^{A^p}(t)$ then $v \in H(t)$ and $t \in H(v)$.

Let us prove that $H^*(t) \subseteq H^{A^p}(t)$. By Conjecture 5 we observe that $H(t) \subseteq E^{A^p}(t)$. Suppose that we have nodes $t, v \in N$ such that $v \in H(t)$ and $t \perp\!\!\!\perp_p v | X$ for some $X \subseteq N \setminus \{t, v\}$. Therefore, it must be that $v \in E^{A^p}(t)$ and $v \notin H^{A^p}(t)$. Thus, $v \in E^{A^p}(t) \setminus H^{A^p}(t)$. By Lemma 4, we have that $t \notin E^{A^p}(v) \setminus H^{A^p}(v)$. By symmetry, $t \notin H^{A^p}(v)$ and thus $t \notin E^{A^p}(v)$. Therefore, $t \notin H(v)$ and Algorithm 2 does not include $t$ and $v$ as neighbors of each other.

□

## 3.2 FINDING THE MARKOV BLANKET

Learning the spouses of the target $t$ is quite similar to the learning the neighbors of $t$. In addition to the data $D$ and target node $t$ we take as input the set $H^*(t)$, learned by Algorithm 2, consisting of the neighbors of $t$. The set $S(t)$ will consist the potential spouses of $t$. As all potential spouses are neighbors of a neighbor of $t$, we need to go through only a subset of all nodes. The set of neighbors $H^*(t)$ is fixed and we keep updating the set $S(t)$. All nodes that have a common child with $t$ and are not neighbors of $t$ are kept in and the rest are discarded. Again, the nodes that remain in $S(t)$ once all nodes have been considered, are considered potential spouses of $t$.

The procedure is summarized in Algorithm 3.

**Algorithm 3** FINDPOTENTIALSPOUSES
**Input:** Data $D$ on node set $N$, a target node $t \in N$, neighbors of the target $H^*(t)$.
**Output:** Potential spouses of $t$.
1: Find set $H'$: the neighbors of the nodes in $H^*(t)$.
2: Initialization: $O \leftarrow H' \setminus (H^*(t) \cup \{t\})$, $S(t) \leftarrow \emptyset$
3: **while** $O$ is nonempty **do**
4:   Choose $v \in O$
5:   $O \leftarrow O \setminus \{v\}$
6:   $Z \leftarrow \{t, v\} \cup H^*(t) \cup S(t)$
7:   $A \leftarrow$ OPTIMALNETWORK($Z, D_Z$)
8:   $S(t) \leftarrow S^A(t)$
9: **end while**
10: **return** $S(t)$

The following lemma shows that the set $S(t)$ will not contain any nodes that are not spouses of the target $t$ in $A^p$.

**Lemma 7.** *Let $S^{A^p}(t)$ be the spouses of the target $t$ in $A^p$. When Assumptions 1 and 2 and Conjecture 5 hold,* Algorithm 3 will return a set $S(t)$ such that $S(t) \subseteq S^{A^p}(t)$.

*Proof.* Let $t$ be the target, $u$ its neighbor and $v$ a neighbor of $u$. A spouse $v$ is added to $S(t)$ only when there is a v-structure $(t, u, v)$ in $A$.

Suppose that there is no v-structure $(t, u, v)$ in $A^p$. Then $t \not\perp\!\!\!\perp_p v | X$ for all $X \subseteq N \setminus \{t, u, v\}$. Now, if a DAG $A$ has the v-structure $(t, u, v)$ then $A_t$ does not contain $u$ and thus $t \not\perp\!\!\!\perp_p v | A_t$. This means that, by local consistency, adding an arc $vt$ to $A$ increases the score. Symmetrically, $A_v$ does not contain $u$ and by local consistency, adding an arc $tv$ increase the score. Since it is always possible to add at least one of these arcs without introducing a cycle, $A$ cannot be the optimal graph.

□

Lemma 7 does not guarantee that all spouses are found (in the limit) using Algorithm 3. Indeed, Figure 1 shows an example where Algorithm 3 leaves one spouse out. Let $t$ be our target. Then, its neighbors are $s$ and $u$. Consider learning an optimal DAG on $\{s, t, u, v\}$. We notice that $t$ and $v$ are not conditionally independent given any subset of $\{s, u\}$ and thus by local consistency the optimal network must contain an arc between $t$ and $v$. Therefore, $v$ cannot be part of a v-structure and is discarded. However, in the original graph $v$ is involved in a v-structure $(t, s, v)$. To find spouses, we use Algorithm 4.

**Algorithm 4** FINDSPOUSES
**Input:** Data $D$ on node set $N$, a target node $t \in N$, neighbors of the target $H^*(t)$, neighbors of the neighbors of the target $H^*(v)$ for all $v \in H^*(t)$.
**Output:** The spouses of $t$.
1: $S^*(t) \leftarrow$ FINDPOTENTIALSPOUSES($D, t, H^*(t)$)
2: **for all** $v \in N \setminus (\{t\} \cup H^*(t))$ **do**
3:   $S(v) \leftarrow$ FINDPOTENTIALSPOUSES($D, v, H^*(v)$)
4:   **if** $t \in S(v)$ **then**
5:     $S^*(t) \leftarrow S^*(t) \cup \{v\}$
6:   **end if**
7: **end for**
8: **return** $S^*(t)$

It is not known whether Algorithm 4 is guaranteed to find all spouses in the limit. Finding a counterexample seems difficult and thus we conjecture as follows.

**Conjecture 8.** *Let $S^{A^p}(t)$ be the spouses of the target $t$ in $A^p$. When Assumptions 1 and 2 hold, Algorithm 4 will return a set $S^*(t)$ such that $S^*(t) = S^{A^p}(t)$.*

Once we have found the neighbors $H^*(t)$ and spouses $S^*(t)$ of a target node $t$, the Markov blanket is simply

$H^*(t) \cup S^*(t)$. The following conjecture that holds if Conjectures 5 and 8 hold summarizes the theoretical guarantees.

**Conjecture 9.** *When Assumptions 1 and 2 hold, the local learning algorithm is optimal in the limit, that is, when the sample size m approaches infinity the algorithm always finds the correct Markov blanket for every target.*

### 3.3 TIME AND SPACE REQUIREMENT

In both Algorithm 1 and 3, the while loop is executed at most $n-1$ times. The time and space requirement inside the loop is dominated by the procedure OPTIMALNETWORK. On a node set $Z$ it runs in $O(|Z|^2 2^{|Z|})$ time and $O(|Z|2^{|Z|})$ space, where in the worst case $|Z| = O(n)$. Thus, these algorithms have a worst case time requirement $O(n^3 2^n)$ and space requirement $O(n 2^n)$. In practice, however, the networks are often relatively sparse and the running times are significantly lower than in the worst case; see the experiments. Algorithms 2 and 4 call Algorithms 1 and 3 at most $n$ times, respectively. Thus, the total time requirement is at most $O(n^4 2^n)$.

The above algorithms were presented for computing the Markov blanket for a single target. If one computes Markov blankets for all nodes, one can store and reuse the the potential neighbor and spouse sets. Thus, in the worst case the Markov blanket for all nodes can be found in $O(n^4 2^n)$ time.

### 3.4 FROM LOCAL TO GLOBAL

Above we have learned Markov blankets of target nodes. Next, we introduce two methods to construct a DAG on the whole node set based on the local results.

The first method uses a constraint-based approach. As mentioned in the previous section, when computing the local neighbor sets we do not need to do the symmetry correction separately for each node. Instead, we can first find the potential neighbors for each node and then get the actual neighbors and build the skeleton using AND-rule; an edge between $u$ and $v$ is added to the skeleton only if both $u$ and $v$ are potential neighbors of each other. Similar way we can also skip the separate symmetry checks when finding the spouses of each node. As a result, we can use the procedure SLL+C described in Algorithm 5 to construct a DAG.

On lines 1–4 of the Algorithm 5 the skeleton $E$ is built and on lines 5–19 the v-structures are directed; this specifies the Markov equivalence class of the DAG. To direct the rest of the arcs which is done on line 20 we can use the rules listed in Pearl (2000). Note that in practice there may be conflicts between v-structures.

**Algorithm 5** SLL+C
**Input:** Data $D$ on node set $N$.
**Output:** Directed acyclic graph $A$.
1: **for all** $v \in N$ **do**
2: $\quad H(v) \leftarrow$ FINDPOTENTIALNEIGHBORS$(D, v)$
3: **end for**
4: $E \leftarrow \{\{u, v\} \mid v \in H(u) \text{ and } u \in H(v)\}$
5: **for all** $v \in N$ **do**
6: $\quad H^*(v) \leftarrow \{u \mid \{v, u\} \in E\}$
7: $\quad S(v) \leftarrow$ FINDPOTENTIALSPOUSES$(D, v, H^*(v))$
8: **end for**
9: $A \leftarrow \emptyset$
10: **for all** $v \in N$ **do**
11: $\quad$ **for all** $u \in S(v)$ **do**
12: $\quad\quad$ **for all** $w$ is a common child of $v$ and $u$ **do**
13: $\quad\quad\quad$ **if** possible without introducing cycles **then**
14: $\quad\quad\quad\quad$ remove $\{v, w\}$ and $\{u, w\}$ from $E$
15: $\quad\quad\quad\quad$ add $vw$ and $uw$ to $A$
16: $\quad\quad\quad$ **end if**
17: $\quad\quad$ **end for**
18: $\quad$ **end for**
19: **end for**
20: direct the rest of the edges without introducing cycles or (if possible) additional v-structures
21: **return** $A$

With a finite sample size, one v-structure could, for example, force an arc to be oriented from $u$ to $v$ and another v-structure from $v$ to $u$.

Another way to construct a DAG from the local results is to use a heuristic such as greedy search with some constraints imposed by the local neighbors. As we will see later in the experiments section, this often leads to a structure with better score compared to the constraint-based approach. Another advantage is that there is no need to find the spouses of the nodes, which is often the more computationally intensive phase in SLL. Algorithm 6 describes SSL+G procedure which uses OR-rule to build a skeleton $E$ of potential edges from the local neighbor sets. Then it calls a greedy search as a subroutine, which may only add an arc if the corresponding edge is present in the skeleton. This approach is similar to the MMHC algorithm by Tsamardinos et al. (2006a) and comes with no correctness guarantees in the limit.

## 4 EXPERIMENTS

We implemented the SLL algorithm as well as both SLL+C and SLL+G algorithms in C++. As OPTIMALNETWORK subroutine we used the dynamic programming algorithm by Silander and Myllymäki (2006) with fallback to greedy equivalence search (GES) (Chickering, 2002) if the number of nodes in the input is larger

**Algorithm 6** SLL+G

**Input:** Data $D$ on node set $N$.
**Output:** Directed acyclic graph $A$.
 1: **for all** $v \in N$ **do**
 2:     $H(v) \leftarrow$ FINDPOTENTIALNEIGHBORS$(D, v)$
 3: **end for**
 4: $E \leftarrow \{\{u, v\} \mid v \in H(u) \text{ or } u \in H(v)\}$
 5: $A \leftarrow$ GREEDYSEARCH$(D, E)$
 6: **return** $A$

| Network | Num vars | max in/out -degree | domain size |
|---|---|---|---|
| ALARM | 37 | 4 / 5 | 2–4 |
| ALARM3 | 111 | 4 / 5 | 2–4 |
| ALARM5 | 185 | 4 / 6 | 2–4 |
| BARLEY | 48 | 4 / 5 | 2–67 |
| CHILD | 20 | 2 / 7 | 2–6 |
| CHILD3 | 60 | 3 / 7 | 2–6 |
| CHILD5 | 100 | 2 / 7 | 2–6 |
| CHILD10 | 200 | 2 / 7 | 2–6 |
| HAILFINDER | 56 | 4 / 16 | 2–11 |
| INSURANCE | 27 | 3 / 7 | 2–5 |
| INSURANCE3 | 81 | 4 / 7 | 2–5 |
| INSURANCE5 | 135 | 5 / 8 | 2–5 |

Table 1: Bayesian networks used in the experiments. Information from Tsamardinos et al. (2006a).

than 20. In the dynamic programming we limited the maximum in-degree of a node to 5. The implementation is available at http://www.cs.helsinki.fi/u/tzniinim/uai2012/. These algorithms were compared against several state-of-the-art algorithms, namely the constraint-based HITON (Aliferis et al., 2010a) for local structure discovery and greedy search (GS), greedy equivalence search (GES) and the max-min hill-climbing (MMHC) (Tsamardinos et al., 2006a) for global structure discovery. The association test used by HITON and MMHC was the built in Assoc function of the implementation, that is, $G^2$ test according to Aliferis et al. (2010a) and Tsamardinos et al. (2006a). For HITON the maximum conditioning set size was set to 5. BDeu prior with equivalent sample size 1 was used in all algorithms when applicable.

In our experiments we used discrete data generated from real world Bayesian networks. We chose a subset of (smaller) data sets which were used by Tsamardinos et al. (2006a) and are freely available online[1]. These are listed in Table 1. Some of the networks have been generated by tiling 3, 5 or 10 copies of the original network using a method by Tsamardinos et al. (2006b). For each network we had 10 independently generated random datasets containing 500, 1000 and 5000 i.i.d. samples.

To accommodate to limited space we chose a representative sample of four of the networks for which we have figures: ALARM5, CHILD10, INSURANCE5, and HAILFINDER. The figures for the rest of the networks are available as an online appendix.

### 4.1 LOCAL LEARNING

In the local learning experiments the goal was to learn both the neighbor sets and the Markov blankets for all nodes. We compared SLL to the constraint-based HITON from Causal Explorer toolkit[2] by Aliferis et al. (2011).

For both the neighbor sets and the Markov blankets we measured the local Hamming distance between the learned set of nodes and the true set of nodes. The Hamming distance of two sets $A$ and $B$ is the number of elements which are contained in only one of the sets, that is, $|A \setminus B| + |B \setminus A|$. For each data set we computed the Sum of Local Hamming Distances over all possible target nodes, call it SLHD.

Figure 2 shows the average SLHDs as a function of the size of the data set for learning the neighbor nodes. As expected, the accuracy of the learned neighbor set improves as the number of data samples increases. In most of the cases SLL beats HITON. The average SLHDs for learning the Markov blankets are shown in Figure 3. The results are similar to those for neighbor nodes.

### 4.2 GLOBAL LEARNING

In the global learning experiments the goal was to learn the entire structure (DAG) of the underlying Bayesian network. We considered both SLL+C and SLL+G heuristics. As the greedy search algorithm we used the steepest ascent hill-climbing with a TABU list of the last 100 structures and a stopping criterion of 15 steps without improvement in the maximum score. These parameters were chosen to be same as in the MMHC and greedy search implementation by Tsamardinos et al. (2006a). We compared our algorithms to greedy search (GS) and max-min hill-climbing (MMHC) from Causal Explorer toolkit as well as greedy equivalent search (GES) from TETRAD software[3].

Some of the algorithms return a DAG but the others return a PDAG, a partially directed acyclic graph which corresponds to a Markov equivalence class. Compared to a DAG, in the corresponding PDAG the edges which are not oriented the same way in all members of the class are undirected. Since the DAGs belonging to a same equivalence class cannot be distinguished using score equivalent scores, we converted all returned DAGs to PDAGs and compared the PDAGs. If the algorithm failed (i.e. did not return a valid DAG) the run was ignored. This only affected GES, which failed to return an acyclic graph in several runs.

---
[1] http://www.dsl-lab.org/supplements/mmhc_paper/mmhc_index.html
[2] http://www.dsl-lab.org/causal_explorer/
[3] http://www.phil.cmu.edu/projects/tetrad/

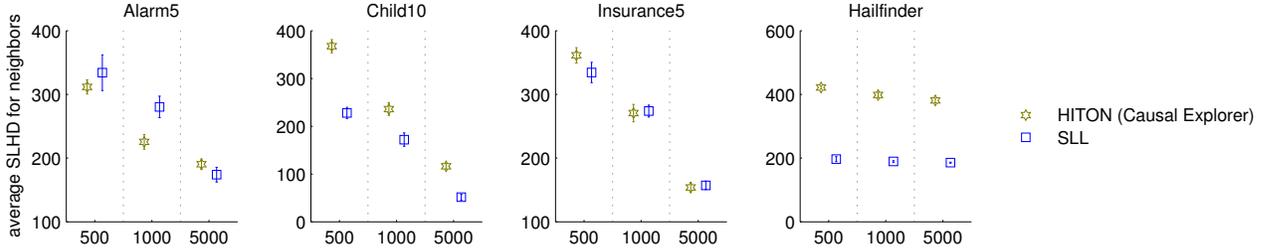

Figure 2: Average SLHD (Sum of Local Hamming Distances) between the returned neighbor sets (parents and children) and the true neighbor sets for different data sizes (500, 1000 and 5000 samples). Standard deviations are shown as vertical bars.

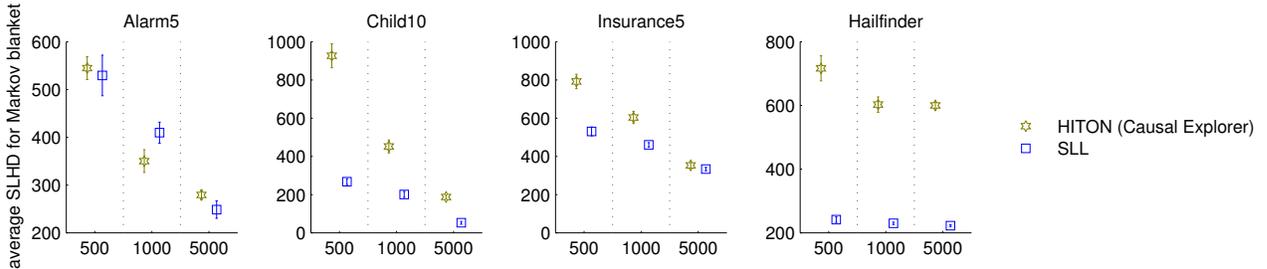

Figure 3: Average SLHD between the returned Markov blankets and the true Markov blankets.

We used two measures the evaluate the goodness of the PDAGs returned by the algorithms: normalized BDeu score and Structural Hamming Distance from the true structure. The normalized BDeu score was obtained by computing the BDeu score with equivalent sample size 1 for a DAG extension of the PDAG and dividing the result by the score of the true structure; the lower the normalized score, the better the structure fits to the data. Average normalized scores are shown in Figure 4. The basic greedy search seems to do surprisingly well, even compared to the GES. For some networks SLL+G works better than MMHC, for other it is the other way around. SLL+C loses to the other methods in most of the cases.

The Structural Hamming Distance (SHD) between two PDAGs is defined as the number of edges which are missing/extra or of wrong type (reversed or directed in one PDAG and undirected in the other). We measured the distances between the resulting PDAGs and the true structures. Figure 5 shows the average SHDs. Compared to the observations for scores, here SLL+C performs clearly better and a basic greedy search much worse.

### 4.3 TIME CONSUMPTION

Figure 6 shows the average running times. For local learning the times include learning both the neighbors and the Markov blankets for all nodes. As in SLL+C the most of the time is spent in local learning part which is the same as running the SLL algorithm for all nodes, these times are combined. The times for HITON are not directly comparable to SLL since it does the symmetry correction for each node separately. In spite of this extra work, HITON is often not any slower than SLL. Also for global learning the rival algorithms are usually faster than SLL+G and SLL+C.

## 5 DISCUSSION

In this paper we have studied prospects of score-based local learning. We have conjectured that the score-based local learning provides same the theoretical guarantees as greedy equivalence search (GES) (Chickering, 2002) and constraint-based local learning (Aliferis et al., 2010a). A natural avenue for future research is to prove these guarantees or lack thereof.

Our experiments suggest that our method provides a competitive alternative for constraint-based local learning. Our algorithm seems to often find local neighborhoods more accurately than the competitor. However, as a downside our algorithm usually consumes significantly more time. We hope that our algorithm could bridge the gap between exact algorithms and constraint-based local learning by providing a way to trade off between accuracy and scalability. The local-to-global approach seem to perform less well compared to various other heuristics. However, in some data sets local-to-global algorithms outperformed all benchmarks in structural Hamming distance especially when there was lots of data. This suggests that with further development, score-based local-to-global heuristic could be competitive in certain types of networks.

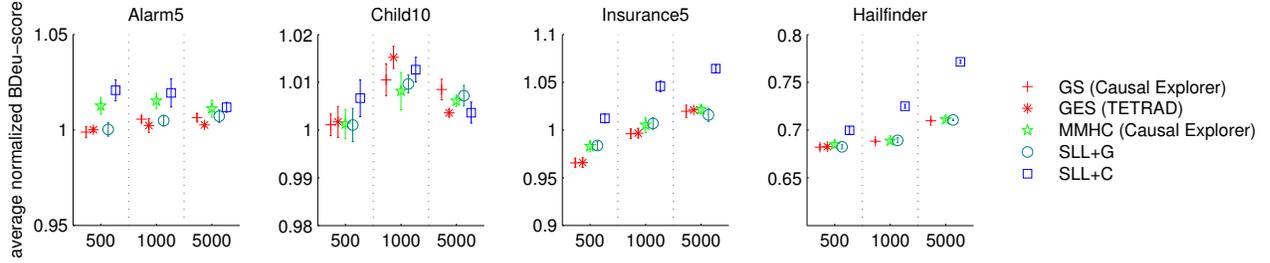

Figure 4: Average normalized score of the returned network structures. Smaller is better and 1.0 corresponds to the true network.

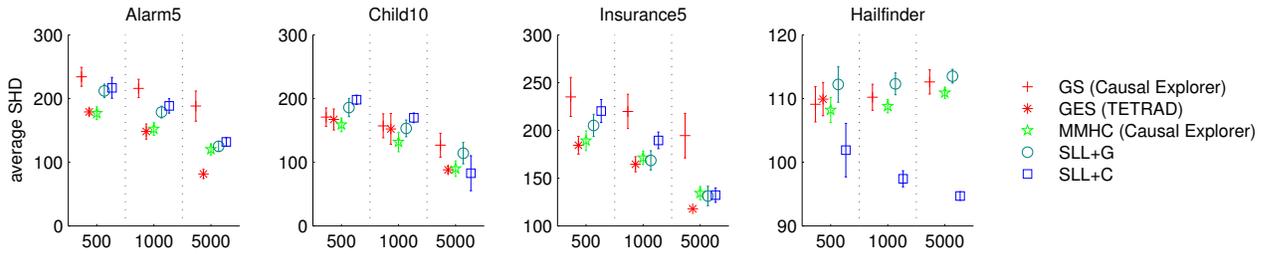

Figure 5: Average SHD (Structural Hamming Distance) between the returned structures and the true structure.

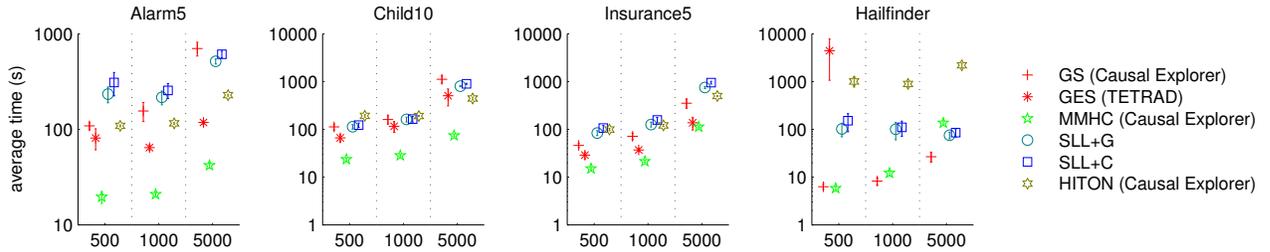

Figure 6: Average running times for the algorithms.

# References

Constantin F. Aliferis, Alexander Statnikov, Ioannis Tsamardinos, Subramani Mani, and Xenofon D. Koutsoukos. Local causal and Markov blanket induction for causal discovery and feature selection for classification part I: Algorithms and empirical evaluation. *Journal of Machine Learning Research*, 11:171–234, 2010a.

Constantin F. Aliferis, Alexander Statnikov, Ioannis Tsamardinos, Subramani Mani, and Xenofon D. Koutsoukos. Local causal and Markov blanket induction for causal discovery and feature selection for classification part II: Analysis and extensions. *Journal of Machine Learning Research*, 11:235–284, 2010b.

Constantin F. Aliferis, Ioannis Tsamardinos, and Alexander Statnikov. Causal explorer: A probabilistic network learning toolkit for discovery. Software, 2011. URL http://discover.mc.vanderbilt.edu/discover/public/causal_explorer/.

David Maxwell Chickering. Learning Bayesian networks is NP-Complete. In Doug Fisher and Hans-J. Lenz, editors, *Learning from Data: Artificial Intelligence and Statistics*, pages 121–130. Springer-Verlag, 1996.

David Maxwell Chickering. Optimal structure identification with greedy search. *Journal of Machine Learning Reseach*, 3:507–554, 2002.

David Maxwell Chickering, David Heckerman, and Chris Meek. Large-sample learning of Bayesian networks is NP-Hard. *Journal of Machine Learning Research*, 5:1287–1330, 2004.

Gregory F. Cooper and Edward Herskovits. A Bayesian method for the induction of probabilistic networks from data. *Machine Learning*, 9(4):309–347, 1992.

James Cussens. Bayesian network learning with cutting planes. In *Proceedings of the 27th Conference on Uncertainty in Artificial Intelligence (UAI)*, pages 153–160. AUAI Press, 2011.

Cassio P. de Campos, Zhi Zeng, and Qiang Ji. Structure learning of Bayesian networks using constraints. In *Proceedings of the 26th International Conference on Machine Learning (ICML)*, pages 113–120. Omnipress, 2009.

David Heckerman, Dan Geiger, and David Maxwell Chickering. Learning Bayesian networks: The combination of knowledge and statistical data. *Machine Learning*, 20(3):197–243, 1995.

Tommi Jaakkola, David Sontag, Amir Globerson, and Marina Meila. Learning Bayesian network structure using LP relaxations. In *Proceedings of the 13th International Conference on Artificial Intelligence and Statistics (AISTATS)*, JMLR: W&CP, pages 358–365, 2010.

Brandon Malone, Changhe Yuan, Eric A. Hansen, and Susan Bridges. Improving the scalability of optimal Bayesian network learning with external-memory frontier breadth-first branch and bound search. In *Proceedings of the 27th Conference on Uncertainty in Artificial Intelligence (UAI)*, pages 479–488. AUAI Press, 2011.

Andreas Nägele, Mathäus Dejori, and Martin Stetter. Bayesian substructure learning - approximate learning of very large network structures. In *European Conference on Machine Learning (ECML)*, pages 238–249, 2007.

Pekka Parviainen and Mikko Koivisto. Exact structure discovery in Bayesian networks with less space. In *Proceedings of the 25th Conference on Uncertainty in Artificial Intelligence (UAI)*, pages 436–443. AUAI Press, 2009.

Judea Pearl. *Causality: Models, Reasoning, and Inference*. Cambridge university Press, 2000.

Tomi Silander and Petri Myllymäki. A simple approach for finding the globally optimal Bayesian network structure. In *Proceedings of the Twenty-Second Annual Conference on Uncertainty in Artificial Intelligence (UAI)*, pages 445–452. AUAI Press, 2006.

Peter Spirtes, Clark Glymour, and Richard Scheines. *Causation, Prediction, and Search*. Springer Verlag, 2000.

Ioannis Tsamardinos, Laura E. Brown, and Constantin F. Aliferis. The max-min hill-climbing Bayesian network structure learning algorithm. *Machine Learning*, 65:31–78, 2006a.

Ioannis Tsamardinos, Alexander Statnikov, Laura E. Brown, and Constantin F. Aliferis. Generating realistic large Bayesian networks by tiling. In *The 19th International FLAIRS Conference*, pages 592–597, 2006b.

Thomas S. Verma and Judea Pearl. Equivalence and synthesis of causal models. In *Proceedings of the 6th Annual Conference on Uncertainty in Artificial Intelligence (UAI)*, pages 255–270. Elsevier, 1990.